\documentclass[lettersize,journal]{IEEEtran}
\usepackage{amsmath,amsfonts}

\usepackage{array}
\usepackage[caption=false,font=normalsize,labelfont=sf,textfont=sf]{subfig}
\usepackage{textcomp}
\usepackage{stfloats}
\usepackage{url}
\usepackage{verbatim}
\usepackage{graphicx}
\usepackage{cite}
\usepackage{xspace}

\newcommand{\norm}[1]{\left\lVert#1\right\rVert}

\usepackage{tikz}
\usepackage{algorithm}
\usepackage[noend]{algpseudocode}
\usepackage[normalem]{ulem}

\usepackage{amsmath,amsfonts,bm}
\usepackage{amssymb}

\usepackage{amsthm}
\theoremstyle{plain}

\newtheorem{assumption}{Assumption}

\newtheorem{corollary}{Corollary}

\newtheorem{theorem}{Theorem}
\newtheorem{definition}{Definition}

\def\1{\bm{1}}

\def\vzero{{\bm{0}}}
\def\vone{{\bm{1}}}

\def\ve{{\bm{e}}}

\def\vp{{\bm{p}}}
\def\vq{{\bm{q}}}

\def\vs{{\bm{s}}}

\def\vu{{\bm{u}}}
\def\vv{{\bm{v}}}

\def\vy{{\bm{y}}}
\def\vz{{\bm{z}}}

\def\mA{{\bm{A}}}

\def\mG{{\bm{G}}}

\def\mI{{\bm{I}}}

\def\mL{{\bm{L}}}
\def\mM{{\bm{M}}}

\def\mP{{\bm{P}}}

\def\mQ{{\bm{Q}}}
\def\mR{{\bm{R}}}
\def\mS{{\bm{S}}}
\def\mT{{\bm{T}}}
\def\mU{{\bm{U}}}
\def\mV{{\bm{V}}}
\def\mW{{\bm{W}}}
\def\mX{{\bm{X}}}

\DeclareMathAlphabet{\mathsfit}{\encodingdefault}{\sfdefault}{m}{sl}
\SetMathAlphabet{\mathsfit}{bold}{\encodingdefault}{\sfdefault}{bx}{n}

\def\gD{{\mathcal{D}}}

\def\gM{{\mathcal{M}}}
\def\gN{{\mathcal{N}}}
\def\gO{{\mathcal{O}}}

\def\gS{{\mathcal{S}}}

\newcommand{\E}{\mathbb{E}}

\newcommand{\R}{\mathbb{R}}

\newcommand{\Var}{\mathrm{Var}}

\DeclareMathOperator{\Tr}{Tr}

\usepackage{hyperref}
\usepackage{enumitem}
\usepackage{subfig}

\begin{document}

\title{Decentralized Differentially Private Power Method}

\author{Andrew Campbell,~\IEEEmembership{Student Member,~IEEE,} Anna Scaglione, ~\IEEEmembership{Fellow, ~IEEE,} Sean Peisert, ~\IEEEmembership{Senior Member, ~IEEE}
        % <-this % stops a space
\thanks{This work was supported in part by the DoD-ARO under Grant W911NF2210228; and in part by the Director, Cybersecurity, Energy Security, and Emergency Response (CESER) Office of U.S. Department of Energy via the Privacy-Preserving, Collective Cyberattack Defense of DERs Project DEAC02-05CH11231.}
\thanks{Andrew Campbell and Anna Scaglione are with Department of Electrical and Computer Engineering, Cornell University, Cornell Tech, New York, 10044 NY, USA (email: ac2458@cornell.edu, as337@cornell.edu)}% <-this % stops a space
\thanks{Sean Peisert is with the Lawrence Berkeley National Laboratory, CA USA (email: sppeisert@lbl.gov)}
}
\markboth{}{}


\maketitle

\begin{abstract}
We propose a novel Decentralized Differentially Private Power Method (D-DP-PM) for performing Principal Component Analysis (PCA) in networked multi-agent settings. Unlike conventional decentralized PCA approaches where each agent accesses the full n-dimensional sample space, we address the challenging scenario where each agent observes only a subset of dimensions through row-wise data partitioning. Our method ensures $(\epsilon,\delta)$-Differential Privacy (DP) while enabling collaborative estimation of global eigenvectors across the network without requiring a central aggregator. We achieve this by having agents share only local embeddings of the current eigenvector iterate, leveraging both the inherent privacy from random initialization and carefully calibrated Gaussian noise additions. We prove that our algorithm satisfies the prescribed $(\epsilon,\delta)$-DP guarantee and establish convergence rates that explicitly characterize the impact of the network topology. Our theoretical analysis, based on linear dynamics and high-dimensional probability theory, provides tight bounds on both privacy and utility. Experiments on real-world datasets demonstrate that D-DP-PM achieves superior privacy-utility tradeoffs compared to naive local DP approaches, with particularly strong performance in moderate privacy regimes ($\epsilon\in[2, 5]$). The method converges rapidly, allowing practitioners to trade iterations for enhanced privacy while maintaining competitive utility.
\end{abstract}


\section{Introduction}
\label{intro}
\IEEEPARstart{T}{he} goal of computing the singular vectors of decentralized data is a critical data analysis task used in both research and commercial settings. For example, the eigenvector estimation of the covariance matrix of a dataset $\mX\in\R^{n\times d}$ can be used for learning distance embeddings, performing a dimensionality reduction, or for the construction of a similarity matrix. However, as the big data paradigm becomes more popular there is a growing need for incorporating privacy into data analysis pipelines. While distributed Principal Component Analysis (PCA) methods \cite{scaglione2008decentralized,wang2023incremental,chai2022practical, froelicher2023scalable,grammenos2020federated,qu2002principal,improved_DPCA} provide some degree of privacy, Differentially Privacy (DP) is required to make formal privacy guarantees. \cite{zari2022membership} demonstrated that the eigenvectors, whether produced in a centralized or decentralized setting, are vulnerable to the membership inference attack unless DP is included. In this paper we focus on the Decentralized Power Method (D-PM) \cite{scaglione2008decentralized} and its adaptions to the DP setting.

\begin{figure}
    \centering
    \includegraphics[width=0.9\linewidth]{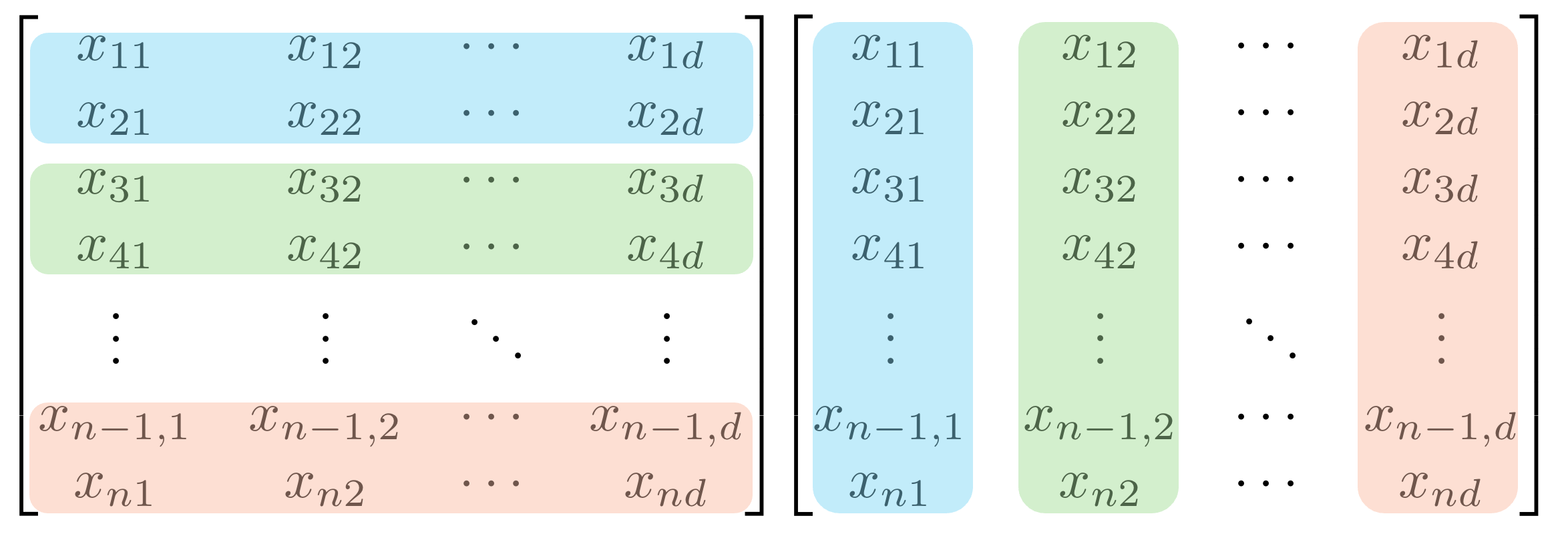}
    \caption{Left: The state split we consider. Right: the standard data split considered by the literature. The color represents a unique agent's dataset.}
    \label{fig:horiz_vs_vert}
\end{figure}

In this work, we aim to estimate the eigenvectors of the covariance matrix $\mX\mX^\top$ in a decentralized manner (or left singular vectors of $\mX$), given a data matrix $\mX \in \mathbb{R}^{n \times d}$. Here, $n$ represents the number of data samples, each of dimension $d$. We assume that our dataset is partitioned row-wise among $m$ agents in a network. Specifically, each agent $i$ holds a data matrix $\mX_i \in \mathbb{R}^{n_i \times d}$ where $\sum_{i=1}^m n_i = n$. 
, such that:
\begin{align}
\mX = \begin{bmatrix}
\mX_1 \\
\vdots \\
\mX_m
\end{bmatrix} \label{eq:our_data_dist}.
\end{align} This is a different set up from the prior art which assumes two different scenarios. The most common one, as in \cite{grammenos2020federated, wang2018differentially, imtiaz2016privacy, ge2018minimax, wang2020principal} distributes the data column wise, i.e each agent has $\mX_i\in\R^{n\times d_i}$ where \begin{align}
    \mX=\begin{bmatrix}
        \mX_1 &\hdots &\mX_m
    \end{bmatrix}. \label{eq_col_par}
\end{align} See Figure \ref{fig:horiz_vs_vert} for a visualization of the two methods.
The second case considered in the literature, as in \cite{nicolas2024differentially,pmlr-v49-balcan16a,guo2021fedpower}, assume each agent has a portion of $\mX\mX^\top$, denoted $\mA_i$ where \begin{align}
    \mX\mX^\top=\sum_{i=1}^m\mA_i.\label{eq_xx_t=A}
\end{align}
This is not possible in our case, since the agents only have access to the block diagonal elements of $\mX\mX^\top$ ($\mX_i\mX_i^\top$). It is worth noting that the partitioning schemes in \eqref{eq:our_data_dist},\eqref{eq_col_par}, and \eqref{eq_xx_t=A} are fundamentally different and require different solutions. To elucidate this, consider the sample wise data splitting in \eqref{eq:our_data_dist}: each agent lacks the full dimensional space of the eigenvectors. Alternatively, consider case \eqref{eq_col_par} or \eqref{eq_xx_t=A}: each agent has the full dimension of the eigenvectors.
 Before continuing we clarify the notation and define some preliminaries. 
 
\subsection{Notation and Preliminaries} $\mX\in\R^{n\times d}$ is the global data matrix with $n$ samples of $d$ dimension. Each of the $m$ agents, denoted by its index $i$, has local dataset $\mX_i \in \mathbb{R}^{n_i \times d}$ where $\sum_{i=1}^m n_i = n$. That is we only assume \eqref{eq:our_data_dist} from here on out. $\mU\bm{\Sigma}\mV^{\top}=\mX$ refers to the Singular Value Decomposition (SVD) of $\mX$ and $\mU_i\bm{\Sigma}_i\mV_i^{\top}=\mX_i$ the SVD of $\mX_i$. Additionally, we always assume the goal is to approximate $\mU$ and not $\mV^\top$. Generally $\bm{\Sigma}$ refers to the covariance of a specific probability distribution, but in the occasion when it refers to the singular values of some matrix $\mA$ we will denote this by $\bm{\Sigma}(\mA)$. Similarly we let $\mU\bm{\Lambda}\mU^\top=\mX$ refer to the eigen decomposition. $\lambda_j$ refers to the $j-$th eigen value of $\mX$. If if is ambiguous as to which eigen value we refer to, we will use the following notation, $\lambda_j(\mA)$ to identify the $j$ eigen value of matrix $\mA$. Eigen values are always assumed to be in descending order, i.e. $\lambda_j\geq\lambda_{j+1}$. Let $\vv$ refer to the principal eigenvector of $\mX\mX^\top$ and let $\vq$ refer to its decentralized power method approximation. As in the case with the matrix decomposition, $\vq_i\in\R^{n_i}$ refers to agent $i$'s portion of the eigenvector where \begin{align}
    \vq=\begin{bmatrix}
        \vq_1\\\vdots\\\vq_m
    \end{bmatrix}.
\end{align}         
In regards to differential privacy we consider the Probabilistic DP definition which is a strictly tighter notion of DP \cite{mcclure2015relaxations}.
The classical $(\epsilon,\delta)-$DP is defined below followed by the $(\epsilon, \delta)-$PDP definition that we use.
\begin{definition}[$(\epsilon, \delta)-DP$ \cite{dwork2014algorithmic}]
    \label{def:classic_dp}
    Let $\gM$ be a randomized algorithm for publishing a query defined over set of databases $\bm{\gD}$. Then $\forall \bm \mX,\bm \mX'\in\bm{\gD}$ such that $\mX$ and $\mX'$ are adjacent, $\gM$ is $(\epsilon,\delta)-$DP if $\forall\gS\subset\operatorname{Range}(\gM)$
    \begin{align}
        \Pr\left[\gM(\mX)\in\gS\right]\leq e^\epsilon\Pr\left[\gM(\mX')\in\gS\right]+\delta. \label{eq:dp}
    \end{align}
\end{definition}
\begin{definition}[$(\epsilon,\delta)-PDP$ \cite{machanavajjhala2008privacy}]
    \label{def:pdp}
    Let $\mX,\mX'\in\mathcal{D}$ be adjacent data sets in database $\mathcal{D}$. Let $\mathcal{M}$ denote a query mechanism over $\mathcal{D}$ with output given by $\tilde{q}$ and where $\tilde{q}\sim f(\tilde{q}|\mX)$. We say $\mathcal{M}$ is PDP if 
    \begin{align}
        \Pr \left(\left|\ln\dfrac{f(\tilde{q}|\mX)}{f(\tilde{q}|\mX')}\right|>\epsilon\right)\leq \delta.
    \end{align}
\end{definition} We chose Definition \ref{def:pdp} as it provides a more intuitive understanding of DP since it connects DP with the hypothesis testing over probability distributions. However, to have a useful notion of privacy we need to formally define the threat model.

\textbf{Threat Model}: We assume that each agent behaves collaboratively—it will not inject false data, ignore received information, collude, etc.—but remains curious and thus potentially attempts to infer the data held by other agents. No agent communicates with the outside world meaning the privacy leakage is only with respect to the agents. More formally, each agent should be prevented from reconstructing the specific data used by other agents for the principal components computation. Furthermore, given that our setup assumes no trustworthy centralized node, it falls under the Local Differentially Privacy (LDP) setting \cite{dwork2006our}. We are interested in providing guarantees on the privacy leakage of node $i$ with respect to (w.r.t) node $j$. That is, given the information node $j$ has available locally and the information node $j$ has received from the network, we want to formally bound node $j$'s ability to infer the private data at node $i$. Finally, we define adjacent datasets by a single row change such that $\norm{\mX-\mX'}\leq 1$. This assumption is standard within the DP PCA literature.
\subsection{Prior Art}
\label{sec:prior_art}
Prior work by \cite{hardt2014noisy} has established analytical convergence guarantees for the Noisy Power Method, thereby providing foundational convergence results for the Differentially Private Power Method (DP-PM). More recently \cite{hardt2014noisy} has been extended to distributed settings where \cite{nicolas2024differentially,guo2021fedpower,wang2020principal,pmlr-v49-balcan16a} proposed federated approaches for the DP-PM algorithm. \cite{nicolas2024differentially,balcan2017differentially,guo2021fedpower} assume a data partition matching \eqref{eq_xx_t=A} while \cite{qu2002principal,grammenos2020federated,wang2018differentially,ge2018minimax} utilize a data partitioning consistent with \eqref{eq_col_par}.
\cite{pmlr-v49-balcan16a} provides an improved convergence rate over \cite{hardt2014noisy} by removing the dependence on $\lambda_{k}-\lambda_{k+1}$ and replaces it by $\lambda_k-\lambda_{q}$ where $k\leq q\leq r$ and where $r$ is the target rank. Additionally, \cite{pmlr-v49-balcan16a} provides a brief extension to the distributed PCA setting providing an algorithm and convergence guarantees.  
In \cite{nicolas2024differentially}, the authors build on the framework in \cite{pmlr-v49-balcan16a} by providing an improved sensitivity analysis, tighter convergence, and better privacy utility trade-off through their usage of Multi-Party Computation (MPC) during aggregation. \cite{nicolas2024differentially} proves DP using the Zero-Concentrated DP method from \cite{bun2016concentrated} and improve the convergence bound from \cite{pmlr-v49-balcan16a} via a more generic sensitivity analysis. On the other hand \cite{guo2021fedpower} bases their work on \cite{hardt2014noisy} and differs further from \cite{nicolas2024differentially} since rather than performing a QR decomposition of the aggregated subspaces, they first align all the subspaces and then perform aggregation and orthogonalization. Finally, \cite{wang2020principal} is analytically built on \cite{hardt2014noisy} but assumes each agent has a single sample. Algorithmically, each agent computes its own sample covariance followed by a noise perturbation and then a central aggregator averages the noisy local sample covariances followed by an eigen decomposition. 
 
We differ from the above works in 3 distinct ways. First we consider a fully-decentralized (multi-agent) scenario. More precisely we do not assume an aggregation or any other other central node and rather assume that all interactions are peer to peer. In fact, to the best of our knowledge we are the first to extend the DP-PM method to the multi-agent setting. Second, again to the best of our knowledge, we are the first to consider the data partition in \eqref{eq:our_data_dist}. The papers that derive their analysis from \cite{hardt2014noisy} assume the partitioning in \eqref{eq_xx_t=A}. Finally we do not base our analysis on \cite{hardt2014noisy}'s seminal work, but rather cleverly construct our algorithm such that we can apply the tools from linear dynamics and high dimensional probability. This allows us to prove convergence from first principles and well established concentration inequalities. Similarly, we appear to be the first to show convergence of the DP-PM utilizing this proof technique.  Below are our contributions. 
\\ 

\noindent\textbf{Contributions}:
\begin{itemize}[align=parleft,nosep,leftmargin=*]\label{contri}
\item We are the first in proposing a D-DP-PM algorithm for estimating the eigenvectors of $\mX\mX^\top$ when $\mX$ is distributed according to \eqref{eq:our_data_dist}. 
\item By design we are able to treat the proposed algorithm as a Gaussian process and provide an $(\epsilon,\delta)-$DP bound utilizing this fact. 
\item We fully describe the dynamics of the proposed algorithm allowing for an error bound that utilizes the Hanson–Wright concentration Inequality \cite{Vershynin_2018}. Additionally we provide an insight into the best choices of the hyper-parameters of the algorithm. 
\item Our convergence analysis shows the effect of the decentralization and the DP noise perturbations. 
\item We provide a quantitative empirical validation on real datasets, both public and private, to verify the efficacy of the proposed algorithm.
\end{itemize}

\section{Proposed Method}\label{sec:alg_intro}
At a high level, our algorithm is a noisy variant of the D-PM algorithm \cite{scaglione2008decentralized}. In this section we provide the complete algorithm and describe the key steps. For clarity and ease of exposition we will first introduce the centralized power method, followed by the decentralized, and finally the proposed Decentralized and Differentially Private Power Method (D-DP-PM).
\subsection{Centralized Power Iteration}
Before introducing the Power Method we first require the following assumption.
\begin{assumption}
    \label{ass:spectrum}
    The eigen values of $\mX\mX^\top$ are distinct and decreasing. That is $\lambda_1>\hdots>\lambda_i>\hdots \lambda_r$, where $r$ is the rank of $\mX\mX^\top$. 
\end{assumption}
Let us consider only the principal eigenvector, i.e. $l=1$, and omit the index. 
The (centralized) PM is an iterative algorithm that computes the dominant eigenvector of a matrix using the following update: \begin{align}
\label{eq:trad_pow}
    \vq^{(t+1)}=\dfrac{\mX\mX^\top \vq^{(t)}}{\norm{\mX\mX^\top \vq^{(t)}}},
\end{align} 
where $\vq^{(0)}$ is a random vector where \begin{align}
    \vq^{(0)}\sim\gN\left(\vzero,\sigma^2_{\vq}\mI_n\right).
\end{align} Running \eqref{eq:trad_pow} for $T$ iterations generates a vector that approximates the principal eigenvector $\mX\mX^\top$. We specify the convergence rate in the following theorem. 
\begin{theorem}[\cite{matrix_comp} Chapter 7.3.1]
\label{th:central}
    If Assumption \ref{ass:spectrum} holds, then the centralized power iteration has the following error bound 
    \begin{align}
        \sin(\vv,\vq^{(T)})\leq \gO\left(\left(\dfrac{\lambda_2}{\lambda_1}\right)^T\right).
    \end{align}
\end{theorem}
Theorem \ref{th:central} indicates that if the eigen gap of $\mX\mX^\top$ is sufficiently large than the power iteration rapidly converges to the true eigenvector. However, in order to generate the whole eigen subspace, we remove the contribution of $\vq^{(T)}$ from the $\mX\mX^\top$ by 
\begin{align}
    (\mX\mX^\top)_{\text{new}}=\mX\mX^\top-\lambda_1\vq\vq^\top,
\end{align}
and repeat \eqref{eq:trad_pow} for the desired rank.
\subsection{Decentralized Power Iteration}
The centralized power method was extended to the decentralized setting \cite{scaglione2008decentralized} where the observation was made that inner-products could be computed in a distributed manner using the average consensus (gossiping) protocol. In order to describe the consensus protocol we first need to introduce some assumptions about the mixing (consensus) matrix $\mW$. 
\begin{assumption}[The mixing matrix]\label{as:mixMatrix}The mixing matrix $\mW$ satisfies the following conditions: 
 \begin{itemize}[nosep,leftmargin=0.2 in]
     \item[i.] $w_{ij}>0$ if and only if there exists an edge between nodes $i$ and $j$.
    \item[ii.] The underlying graph is undirected. %, implying that $\mW$ is symmetric.
     \item[iii.] $\mW$ is doubly-stochastic.
     \item[iv.] The agent network, $\mathcal{G}(\{1,\cdots,m\},E)$ is strongly connected.
     %, implying that $\mW$ is irreducible.
\end{itemize}
\end{assumption}
A consequence of Assumption \ref{as:mixMatrix} is that $0\leq\lambda_2(\mW)<1$.

Therefore, for mixing matrix $\mW$ (satisfying Assumption \ref{as:mixMatrix}) and for consensus steps $c$, the inner product at node $i$ is: 
\begin{align}
\mX^\top\vq+\ve&=m\sum_{j=1}^m\left(\mW^c\right)_{ij}\mX_j^\top\vq_j \label{eq:con_as_inner}\\
\norm{\ve}&=\gO\left(\lambda^c_2\left(\mW\right)\right)\nonumber.
\end{align} A brief proof of this fact is provided in Appendix \ref{proof:cons}.
The upshot of \eqref{eq:con_as_inner} is that, via consensus aggregation, agents can approximate the projection over the other agents data, and if $c$ is sufficiently large, then error due to consensus tends to 0. A critical note here is that each agent only needs to compute its own $\mX_i^\top\vq_i$ and recall that $\vq_i$ is a $n_i$ dimensional vector not an $n$-dimensional one. That is each agent only needs to track and share its portion of $\vq$. However, at the last iteration, to ensure that each agent has the entire $\vq$ agents must share their local $\vq_i$ with the entire network.   
\subsection{Decentralized DP-Power Iteration:}
The above discussion on the PM and its mulit-agent variant come from the existing literature. However, the addition of DP into the decentralized PM is one of our core contributions. See Algorithm \ref{alg:cov_approx_power} for the complete method. We provide the details below. 
As stated in the previous section, each agent in the decentralized PM algorithm only shares the projected vector $\vz_i:=\mX^\top_{i}\vq_i^{(t)}$ (step 6 Algorithm \ref{alg:cov_approx_power}) via consensus. Crucially, since each agent communicates only its own segment of the projected eigenvector, no agent ever tracks the full vector $\vq$ thus improving the privacy-utility trade off.
This fact, together with the independent and random initialization of each agent's vector $\vq_i^{(0)}$, inherently provides additional privacy that we deliberately utilize. 
%By design $\vq_i^{(0)}\sim \mathcal{N}(\vzero,\sigma_{\vq}^2\mI)$.
Note that in our setting we assumed each $\vq^{(0)}_i$ has the same covariance, $\sigma_{\vq}^2\mI$. While not a requirement for convergence or privacy, we   
chose this to simplify exposition and analysis. Furthermore, we chose Gaussian noise since if we preserve Gaussianity, then we can explicitly write the distribution of each agent's stacked release. 
Continuing in this vein, rather than normalizing $\vq^{(t)}$ by its norm, we use a predetermined scalar $\alpha$, thus preserving Gaussianity and saving on privacy. This is critical for the tractability of the analysis since it allows us to stack the data releases of every agent into a single multi-variate Gaussian observation.
Finally, at each iteration of the algorithm we add $\vp_i^{(t)}\sim\gN(\left(\vzero,\sigma_{\vp}(t)^2\mI_n\right)$ after consensus, projection, and rescaling has occurred (step 7 Algorithm \ref{alg:cov_approx_power}). We add the per iteration noise post consensus aggregation because the first iteration is private by the initial random hidden state $\vq^{(0)}$ and because it simplifies the analysis since the noise matches the dimensionality of the release $\vq^{(T)}$. 
 
However, to complete the eigenvector computation each agent must share its own $\vq^{(T)}_i$ with all agents (step 8 of Algorithm \ref{alg:cov_approx_power}). Note, that we do not need to include any additional noise for privacy since step 7 has already noised the local $\vq_i$. Finally, in order to compute the remaining eigenvectors we sequentially remove the contribution from the dominant eigenvector and repeat the algorithm. Concretely, where 
$\vz_i^{(t+1/2)}:=m\sum^{n}_{j}\left(\mW^c\right)_{ij}\vz_j^{(t)}$ (step 6 Algorithm \ref{alg:cov_approx_power}), each agent updates its local dataset by 
\begin{align}
    \mX^{}_i&=\mX_i-\vq_{l-1}\vz^{(T+1/2)}_i\nonumber
    %&=\mX_i-\vq_{l-1}\mX^\top\vq_{l-1} + \vq_{l-1}\gO\left(\lambda^c_2\left(\mW\right)\right),
\end{align}
which projects the data orthogonally to the directions of the leading singular vector. 
% Then the D-DP-PM can be re-run to generate the next singular vector and so on. 
See Algorithm \ref{alg:cov_approx_power} for the full details. 
\begin{algorithm}
    \caption{D-DP-PM}
    \label{alg:cov_approx_power}
    \begin{algorithmic}[1]
    \State {\bfseries Init:\nonumber} $\mX_i$, rank $r$, $c$ consensus steps, scaling factor $\alpha$, $\vq^{(0)}_{il}\sim\gN\left(\vzero,\sigma_q^2\mI\right)$,  $\vp_i^{(t)}\sim\gN\left(\vzero,\sigma_p^2(t)\mI\right)$%, and $\vq_{-1}=\vzero$. 
    \For{$l\in[1,r]$}
    \State{$\mX_i=\mX_i-\vq_{l-1}\vz^{(T-1/2)}$}
    \For{$t\leq T$}
    \State{$\vz_i^{(t)}=\mX_i^\top\vq^{(t-1)}_{i,l}$}
    \State{$\vz_i^{(t+1/2)}=m\sum^{n}_{j}\left(\mW^c\right)_{ij}\vz_j^{(t)}$}
    \State{$\vq_{il}^{(t)}=\alpha\mX_i\vz_i^{(t+1/2)}+\vp_i^{(t)}$}
    \EndFor

    \State{Share $\vq_{il}^{(T)}$, Receive $\vq_{jl}^{(T)}$}
    \State{$\vq_{l}^{(T)}=\dfrac{\vq_{l}^{(T)}}{\norm{\vq_{l}^{(T)}}}$}

    \EndFor
    \State{return $\hat{\mU}^{(r)}=\begin{bmatrix}
        \vq^{(T)}_1\hdots \vq_r^{(T)}
    \end{bmatrix}$}
    \end{algorithmic}
\end{algorithm}

\section{Analysis}
\label{sec:analysis}
We break the analysis up into two sections, the first analyzes the privacy aspects of Algorithm \ref{alg:cov_approx_power} and the second focuses on the convergence. 
\subsection{Privacy}
In this section, we introduce Theorem \ref{th:dp}, which provides conditions on the algorithm's parameters to ensure it satisfies $(\epsilon, \delta)$-DP. 
%Before continuing we want to reiterate the algorithm design decisions that enhance the privacy-utility trade off. The    
The key fact, which we noted in the previous section, is that the initial noise distribution and, the release of each $\vz_i$, and $\vq_i^{(T)}$ are all Gaussian. Therefore, if we vertically stack all the releases across the entire network, we notice that this vector is a multivariate Gaussian. That is,
\begin{align}
    \vy^{(T)}&\in\R^{(mdT+n)}:=\nonumber\\
    &\begin{bmatrix}
        \vz_1^{(1)}&\hdots &\vz_m^{(1)}&\vz_1^{(2)}&\hdots &\vz_m^{(T)}&\vq^{(T)}
    \end{bmatrix}^\top.
\end{align} In fact, because we know the dynamics of the entire algorithm, we can explicitly write out $\vy^{(T)}$. Which we provide in the following section.
\subsubsection{Explicit form of the network release} For the sake of analysis we want to look at the algorithm with respect to what the network as a whole observes. In order to continue we must define the following notation. Let 
\begin{align}
    \operatorname{D}\left(\mX\right):=\begin{bmatrix}
        \mX_1 & \hdots & \vzero\\
        \vdots & \ddots & \vdots\\
        \vzero & \hdots & \mX_m
    \end{bmatrix},\nonumber
\end{align}
and 
\begin{align}
    \bm{\Xi}&:=\operatorname{D}(\mX)\left(m(\mW^c\otimes \mI_d)\right)\operatorname{D}^\top(\mX).\label{eq_xi}
\end{align} 
It is worth noting that $\bm{\Xi}$ is the consensus version of the $\mX\mX^\top$. Given this observation and the technique in Appendix \ref{proof:cons} we can show that 
\begin{align}
    \norm{\bm{\Xi}-\mX\mX^\top}=\gO\left(\lambda_2^c(\mW)\right).
\end{align}
Additionally we define $\mP$ to be the stacked noise such that 
\begin{align}
    \mP:&=\begin{bmatrix}
        \vp^{(1)}\\
        \vdots \\
        \vp^{(T)}
    \end{bmatrix} \qquad \mP\sim\mathcal{N}(\vzero, \Sigma_{\mP})\nonumber \\
    \Sigma_{\mP}:&=\operatorname{diag}(\sigma_{\vp}^2(1),\hdots, \sigma_{\vp}^2(T))\otimes \mI_n.
\end{align}
Using the above we can write the update steps of the algorithm over the entire network as 
\begin{align}
    \vz^{(t)}&=\operatorname{D}^\top\left(\mX\right)\vq^{(t-1)}\nonumber\\
    \vz^{(t+1/2)}&=\left(\mW^c\otimes \mI_d\right)\vz^{(t)}\nonumber\\
    \vq^{(t)}&=\alpha\operatorname{D}\left(\mX\right)\vz^{(t+1/2)}+\vp^{(t)},\nonumber\\
\end{align}
which we can further rewrite in terms of the initial noise $\vq^{(0)}$ and the per iteration noise $\vp^{(t)}$ yielding 
\begin{align}
\vz^{(t)}&=\operatorname{D}^\top(\mX)\left(\alpha\bm{\Xi}\right)^{t-1}\vq^{(0)}+ \operatorname{D}^\top(\mX)\sum^{t-1}_{k=1}\left(\alpha \bm{\Xi}\right)^{t-1-k}\vp^{(k)}\label{eq:z}\\
\vq^{(t)}&=\left(\alpha\bm{\Xi}\right)^{t}\vq^{(0)}+ \sum^{t}_{k=1}\left(\alpha \bm{\Xi}\right)^{t-k}\vp^{(k)}\label{eq:q}.
\end{align}
Using \eqref{eq:z} and \eqref{eq:q} we can write, where $\mM\in\R^{(mdT+n)\times n}$ and $\mL\in\R^{(mdT+n)\times nT}$, 
\begin{align}
    \vy^{(T)}=\mM\vq^{(0)}+\mL\mP.
\end{align}
Explicitly $\mM$ is given by, where $(\mM)_t$ denotes the $t-$th block,  
\begin{align}
    (\mM)_t=\begin{cases}
        \operatorname{D}^\top(\mX)(\alpha\bm{\Xi})^{(t-1)} & t\leq T\\
        \left(\alpha\bm{\Xi}\right)^{t} & t=T+1
    \end{cases}.
\end{align}
We note that the first $T$ blocks of $\mM$ correspond to the $\vz$ releases while the last block corresponds to the final $\vq^{(T)}$ release. In a similar fashion, we note that $\mL$ is a lower triangular matrix responsible for summing the $\vp^{(t)}$ noise terms. If we let $(\mL)_{kt}$ refer to the $k$-th row and $t$-th column block, then $\mL$ is defined by 
\begin{align}
    (\mL)_{tk}:=\begin{cases}
        \vzero, & \text{if } t=1 \text{ or } k>t\\
        \operatorname{D}^\top(\mX) & \text{if } t>1 \text{ and } t=k\\
        \operatorname{D}^{T}(\mX)(\alpha\bm{\Xi})^{t-1-k}, & \text{if }1<k<t
    \end{cases}
\end{align}

While it is useful to write down the entire network release we are ultimately interested in the release of an individual agent since the PDP definition is equivalent to bounding the individual privacy loss of each agent. More formally, we want to bound node $j$'s ability to infer node $i$'s private information. To do this we define $\vy_i$ as the stacked releases of agent $i$. Formally, we define the following selection matrices, where $r_1,\hdots,r_{n_i}$ refer to the global indices of agent $i$'s portion of $\vq$. 
\begin{align}
    \mT_i&:=\mI_T\otimes e_i^\top\otimes \mI_d\\
    \mR_i&:=\begin{bmatrix}
        e^{\top}_{r_1}\\
        \vdots \\
        e^{\top}_{r_{n_i}}
    \end{bmatrix}\\
    \mS_i&:=\begin{bmatrix}
        \mT_i & \vzero \\
        \vzero & \mR_i
    \end{bmatrix}.
\end{align}
Therefore 
\begin{align}
    \vy_i^{(T)}:&= \mS_i\mM\vq^{(0)}+\mS_i\mL\mP\nonumber \\
    :&=\mM_i\vq^{(0)}+\mL_i\mP
\end{align}
which nicely yields that 
\begin{align}
\vy_i^{(T)}\sim\mathcal{N}\left(\vzero,\mM_i\Sigma_{\vq}\mM_i^\top+\mL_i\Sigma_{\mP}\mL_i^\top\right)
\end{align}
However, this is not sufficient for bounding the privacy leakage of node $j$ with respect to node $i$. Node $i$ knows its own internal randomness and has received in an iterative fashion information about node $j$ that has been incorporated into its local variables. To account for this, node $i$ has to condition on its local randomness and treat its received data as a random variable. Formally, we want to find the distribution of $\vy_i^{(T)}|\vq_i,\vp_i^{(1)},\hdots,\vp_i^{(T)}$. In order to get this distribution we need to further split $\mM_i$ and $\mL_i$. Let $\mM_i^{u}$ refer to what agent $i$ has and let $\mM_i^{-u}$ refer to what agent $i$ receives from its neighbors. That is we can write 
\begin{align}
    \mM\vq^{(0)}&=\mM_i^{u}\vq_i^{0}+\mM_i^{-u}\vq_{-i}^{(0)}\\
    \mL\mP&=\mL_{i}^{u}\mP_i+\mL_i^{-u}\mP_{-i},
\end{align}
which yields
\begin{align}
    &\vy_i^{(T)}|\vq_i,\vp_i^{(1)},\hdots,\vp_i^{(T)}\nonumber\\&~~\sim \mathcal{N}\left(\mM^{u}_i\vq^{(0)}_i+\mL^u_{i}\mP,\right.\nonumber\\&~~~~~~~~~~~~\left.\mM^{-u}_{i}\Sigma_{\vq_{-i}}\left(\mM^{-u}_i\right)^\top+ \mL^{-u}_{i}\Sigma_{\mP_{-i}}\left(\mL^{-u}_{i}\right)^\top\right).\label{eq:y_i_cond}
\end{align}
Armed with \eqref{eq:y_i_cond} we can formally bound the privacy leakage of node $j$ by bounding the leakage of agent $j$ with respect to agent $i$ and taking the max of over all $i$. Proceeding in the fashion we have the following theorem.
% \begin{align}
%     \mB_i:&=\E\left[\mZ_i^{(T-1)}\right]\\
%     \mC_i:&=\mE\left[\left(\mZ_i^{(T-1)}-\mB_i\right)\left(\mZ_i^{(T-1)}-\mB_i\right)^\top\right],
% \end{align}
% $\mB_i:=\E\left[\mZ_i^{(T-1)}\right]$ and 
% $\mC_i:=\mE\left[\left(\mZ_i^{(T-1)}-\mB_i\right)\left(\mZ_i^{(T-1)}-\mB_i\right)^\top\right]$ 
% as the mean and covariance matrix at agent $i$ after $T-1$ iterations. We denote the adjacent means and covariances by $\mB'_i$ and $\mC_i'$. Furthermore, where $\Delta_{\vq}$ is the sensitivity of $\vq$ as defined in \cite{nicolas2024differentially}, we define $\sigma_{\vu}:=\Delta_{\vq}\sqrt{2\ln(1.25/\delta_1)}\epsilon_1^{-1}$ where $\sigma_{\vu}^2$ is the variance of the noise distribution in step 9 Alg. \ref{alg:cov_approx_power}. 

\begin{theorem}%[Lemma 1 of \citep{ramakrishna2023differential}]
\label{th:dp}
    For any $\beta>0$ and where $\operatorname{D}_{\beta+1}\left(\vy_i^{(T)}||{\vy_i^{(T)}}'\right)$ refers to the R\'enyi-Divergence between the distribution of $\vy^{(T)}|\mX,\vq_i,\vp_i^{(1)},\hdots,\vp_i^{(T)}$ and $\vy^{(T)}|\mX',\vq_i,\vp_i^{(1)},\hdots,\vp_i^{(T)}$ respectively. Then  
    \begin{align}
        \delta_i&:=\Pr\left(\ln\frac{f\left(\vy^{(T)}_i|\mX,\vq_i,\vp_i^{(1)},\hdots,\vp_i^{(T)}\right)}{f\left(\vy_i^{(T)}|\mX',\vq_i,\vp_i^{(1)},\hdots,\vp_i^{(T)}\right)}>\epsilon\right)\nonumber\\&\leq \inf_{\beta}\exp{\left(-\beta\epsilon + \beta \operatorname{D}_{\beta+1}\left(\vy^{(T)}_i||{\vy^{(T)}_i}'\right) \right)}
    \end{align}
    We perform a relabeling of the mean and covariance of \eqref{eq:y_i_cond} yielding
    \begin{align}
        \vy_i^{(T)}|\mX,\vq_i,\vp_i^{(1)},\hdots,\vp_i^{(T)}&\sim \mathcal{N}\left(\mu_{\vy_i},\Sigma_{\vy_i}\right)\\
        {\vy_i^{(T)}}'|\mX',\vq_i,\vp_i^{(1)},\hdots,\vp_i^{(T)}&\sim \mathcal{N}\left({\mu_{\vy_i}}',{\Sigma_{\vy_i}}'\right),
    \end{align}
    and by applying \cite{gil2011renyi}, $\operatorname{D}_{\beta+1}\left(\vy^{(T)}_i||{\vy^{(T)}_i}'\right)$ has a closed form expression given by
    \begin{align}
        % &\operatorname{D}_{\beta+1}\left(\vy^{(T)}_i||{\vy^{(T)}_i}'\right)\nonumber \\
        &\mathbf A_{\beta+1} := ({\beta+1})\,\Sigma_{\vy_i}^{-1} \;+\; (-{\beta})\,({\Sigma_{\vy_i}}')^{-1}, \\[6pt]
&D_{\beta+1}(P\|Q)\nonumber\\
  &~~~= \frac{1}{2}\Biggl[
      {(\beta+1)}(\boldsymbol\mu_{\vy_i}-\boldsymbol{\mu_{\vy_i}}')^{\top}
      \mathbf A_{\beta+1}^{-1}
      (\boldsymbol\mu_{\vy_i}-\boldsymbol{\mu_{\vy_i}}')
      \;-\;\nonumber\\
      &~~~~~~~~~~~~~\frac{1}{{\beta}}\,
      \ln\!\left(
        \frac{\det\mathbf A_{\beta+1}^{-1}}
             {\det({\Sigma_{\vy_i}})^{\,{-\beta}}\,
              \det({{\Sigma_{\vy_i}}'})^{\,{\beta+1}}}
      \right)
    \Biggr].
    \end{align}
\end{theorem}
A short proof for Theorem \ref{th:dp} is provided in Appendix \ref{proof:dp}. The take away from Theorem \ref{th:dp} is that, because we designed Algorithm \ref{alg:cov_approx_power} to preserve Gaussianity, we can explicitly write the distribution at each agent and then use the well established tools to provide a privacy bound.   
% However, to understand the asymptotic behavior we introduce the following corollary. 
% \begin{corollary}[Corollary 2 of \cite{ramakrishna2023differential}]\label{cor:dp} Assuming that $s\gg \max\left(1,\lambda_1(\bm{\Gamma}_i)\right)$, then \begin{align}
%         \delta_{i}\leq \exp\left\{1/2s\left(\bm{\mu}_i^\top\bm{\Gamma}_i\bm{\mu}_i-\ln{\det{\bm{\Gamma}_i}}\right)\right\}e^{-\frac{\epsilon}{s}}.\label{eq_cher}
%     \end{align}
% \end{corollary}
% The proof for Theorem \ref{th:dp} can be found in the appendix along with an explicit formulation of the Gaussian  $\mZ^{(T-1)}$ in terms of the parameters, $\alpha$ ,$\mW$, and $\mX$, which can be found in Appendix \ref{app:chernoff}. \\ 
     
\subsection{Convergence}
Before continuing we need to introduce some additional assumptions about the distributed data $\mX$ and the effect of the convergence error.
\begin{assumption}
\label{ass:E_bound}
    Suppose
    \begin{align}
        E:= \norm{\bm{\Xi}-\mX\mX^\top},
    \end{align} then we assume that 
    \begin{align}
        E\leq \lambda_1-\lambda_2.
    \end{align}
    
\end{assumption}

To provide bounds on the final error in $\vq^{(T)}$ we utilize the a similar strategy as the DP proof. Recall by \eqref{eq:q} we can rewrite $\vq^{(T)}$ in terms of linear combinations of the initial and per iteration noise meaning we can full specify the final distribution of $\vq^{(T)}$. The idea is that we can use a well understood tail bound on a Gaussian vector. More formally, we consider our error metric to be, where $\vv$ is the true principal eigenvector of $\mX\mX^\top$, 
\begin{align}
    \sin(\vv,\vq)=\dfrac{\norm{(\mI-\vv\vv^\top){\vq}}}{\norm{{\vq}}}. \label{eq:err_metric}
\end{align}
The core idea for bounding \eqref{eq:err_metric} is to split the error into two components. The first, is measuring how close the version of $\mX\mX^\top$ computed through the ``consensus" iterations, denoted as $\bm{\Xi}$, is to $\mX\mX^\top$. The second, is how close the D-DP-PM gets to the principal eigenvector of $\bm{\Xi}$. Then, by applying Davis-Khan \cite{davis1970rotation} and Hanson–Wright \cite{Vershynin_2018} concentration inequality we can, with high probability bound $\sin(\vv,\vq)$. 

This approach is one of our contributions since it completely deviates from the state-of-the-art\cite{guo2021fedpower,wang2018differentially, pmlr-v49-balcan16a,hardt2014noisy}. Because \cite{guo2021fedpower,wang2018differentially, pmlr-v49-balcan16a} are extensions/modifications of the technique developed by \cite{hardt2014noisy}, we will only highlight the key differences between our work and \cite{hardt2014noisy}. One immediate observation is that \cite{hardt2014noisy} does not preserve Gaussianity and thus cannot utilize our approach. \cite{hardt2014noisy} alternatively uses a high probability bound on the per iteration noise to verify that it is bounded such that the size of the perturbation does not push the iterate by a greater amount than the PM convergence. While both methods are valid we have one distinct advantage with our approach. Our proof comes directly from first principles meaning the logic is straight forward and the analytical tools are well understood leading to clean and clear analysis. Below is the statement of the convergence theorem.

\begin{theorem}
    \label{th:conv}
    Let $\vq^{(T)}$ be the final unnormalized eigenvector produced by Algorithm \ref{alg:cov_approx_power} such that $\mW$ and $\mX$ satisfy Assumptions \ref{ass:spectrum}-\ref{ass:E_bound}, where 
    \begin{align}
        \vq^{(T)}\sim\gN\left(\vzero,\bm{\Omega}\right).
    \end{align} Furthermore, define $\mP:=(\mI-\vv\vv^\top)$ and $\mQ:=\bm{\Omega}^{1/2}\mP\bm{\Omega}^{1/2}$ with $\Theta:=1-\dfrac{\vv^\top\vv}{\Tr(\Omega)}+\Delta$ where $\Delta \geq \dfrac{2\norm{ \left(\mQ-\Theta\bm{\bm{\Omega}}\right)}_F\sqrt{\log(1/\gamma)}+2\norm{ \left(\mQ-\Theta\bm{\bm{\Omega}}\right)}_2\log(1/\gamma)}{\Tr(\bm{\Omega})}.$ Finally, if 
    $\mu_i$ denotes the $i-$th eigen value of $\bm{\Xi}$ and  
\begin{align}
    \rho:=\left(\dfrac{\sigma_\vq^2+\sum_{k=1}^{T}(\alpha\mu_2)^{-2k}\sigma_\vp^2(k)}{\sigma_\vq^2+\sum_{k=1}^{T}(\alpha\mu_1)^{-2k}\sigma_\vp^2(k)}\right), 
\end{align}
then \begin{align}
    &\dfrac{\norm{(\mI-\vv\vv^\top)\vq^{(T)}}^2}{\norm{\vq^{(T)}}^2}\leq\nonumber \\
      &~~~~~\dfrac{2\norm{ \left(\mQ-\Theta\bm{\bm{\Omega}}\right)}_F\sqrt{\log(1/\gamma)}+2\norm{ \left(\mQ-\Theta\bm{\bm{\Omega}}\right)}_2\log(1/\gamma)}{\Tr(\bm{\Omega})} \nonumber\\
      &~~~~~+2\left(\dfrac{n_im\lambda_2^c(\mW)}{\lambda_1-\lambda_2}+(n-1)\rho\left(\dfrac{\mu_2}{\mu_1}\right)^{2T}\right)\nonumber \\
    &\qquad\text{w.p  } 1-\gamma. \label{eq_final}
\end{align}
\end{theorem}
The full proof for Theorem \ref{th:conv} is in Appendix \ref{proof:conv}. However, the astute reader will notice that we have not provided any indication on an appropriate choice of the parameters $\sigma_{\vq}$, $\alpha$, or $\sigma^2_{\vp}(t)$. In fact, one of the advantages of our analysis is that it provides insight on the best choice of parameters for convergence through the following corollary.
\begin{corollary}
    \label{cor:params} If $\dfrac{1}{\mu_1}<\alpha<\dfrac{1}{\mu_2}$ and $\sigma_{\vp}(t)\leq \left(\dfrac{\mu_2}{\mu_1}\right)^{t}$, then \begin{align}
        \rho\leq 1+ \dfrac{1}{\sigma^2_{\vq}(1-(\alpha\mu_1)^{-2})},
    \end{align}
    which implies that $\rho$ is bounded by a constant.
\end{corollary}
For a proof of Corollary \ref{cor:params} see Appendix \ref{proof:params}. While Theorem \ref{th:conv} and Corollary \ref{cor:params} suggest a larger $\sigma_{\vq}$, this has to be balanced by the choice of $\alpha$. However, a reasonable choice $\sigma_\vq$ is the one which keeps the norm of $\vq$ as close to unit. Therefore, by applying the Laurent–Massart inequality \cite{laurent2000adaptive} we get that $\sigma_\vq=\dfrac{1}{\sqrt{n}}$. 

Theorem \ref{th:conv} in conjunction with Corollary \ref{cor:params} directly connects the consensus aggregation error, the Gaussian perturbations, and the convergence of the power method. Theorem \ref{th:conv} indicates that the proposed D-DP-PM method recovers the expected convergence rate as in the centralized case, but retains an additive error from the consensus aggregation and the DP noise additions. Furthermore, in the centralized case the error is bounded by the ratio of $\frac{\lambda_2}{\lambda_1}$ while the decentralized case is with respect to $\frac{\mu_2}{\mu_1}$. Intuitively and formally this is clear since in the decentralized setting the power iteration is with respect to the network matrix $\bm{\Xi}$ not $\mX\mX^\top$.

We note that the convergence theorem indicates the impact of the network topology and consensus errors via the $\lambda_2^c(\mW)$ term. An additional, convenient, observation is that the term $\lambda_2^c(\mW)$ can be made arbitrarily small by choosing $c$ to be very large. In fact, additional rounds of consensus, for a given iteration, do not increase the privacy leakage since the points shared during consensus have already been made public and thus no additional private information is included. 

The key take away is that the D-DP-PM will converge towards $\vv$ at a similar rate to the centralized case, but will reach a saturation point where it can not improve due to the consensus errors and the DP noise. However, given that the algorithm designer has choice over $\sigma^2_{\vp}(t)$ and $c$, this saturation point can be significantly reduced. 
%\subsection{Comparison to SOTA.}

\section{Empirical Evaluation}
In this section we describe the experimentation performed and analysis of the results. We compare our algorithm to the naive approach, which we denote as LDP, of adding Gaussian noise locally to each agent's dataset followed by the sharing of its perturbed data. Then each agent can perform a local SVD over the noisy data to generate the eigenvector estimation. We do not provide a comparison with the recent works of \cite{pmlr-v49-balcan16a,nicolas2024differentially,guo2021fedpower,wang2020principal} since they all consider a different problem. The above works either consider a data distribution given by \eqref{eq_col_par} or \eqref{eq_xx_t=A}. Furthermore, we are the only work that considers the multi-agent set up as opposed to the classic federated or centralized processing node approach. Because of these two facts an empirical comparison with the state-of-the-art would be uninformative.
\begin{figure*}[!ht]
\centering
\subfloat[]{\includegraphics[width=0.42\textwidth]{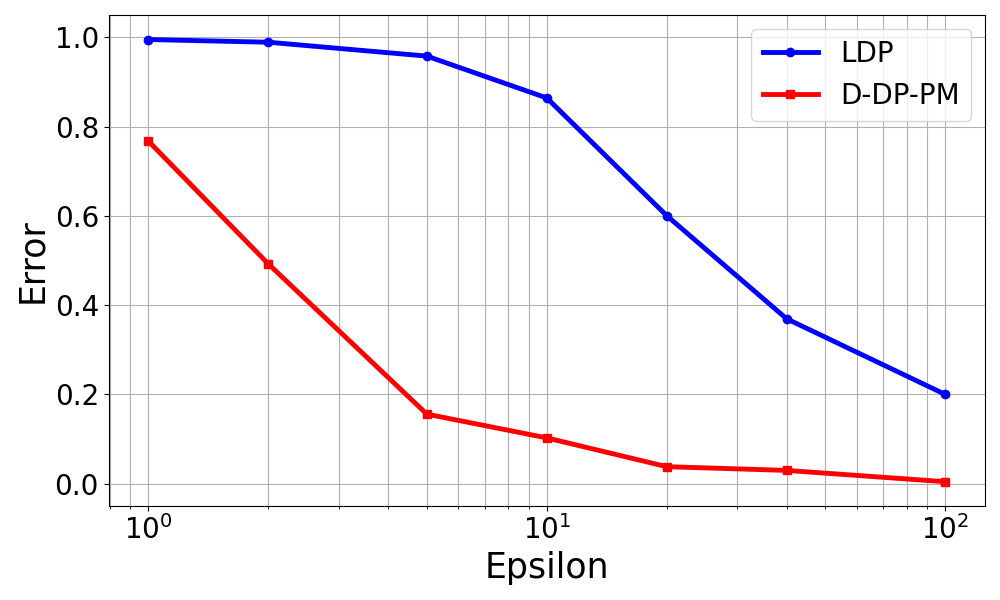}}
\hfill
\subfloat[]{\includegraphics[width=0.42\textwidth]{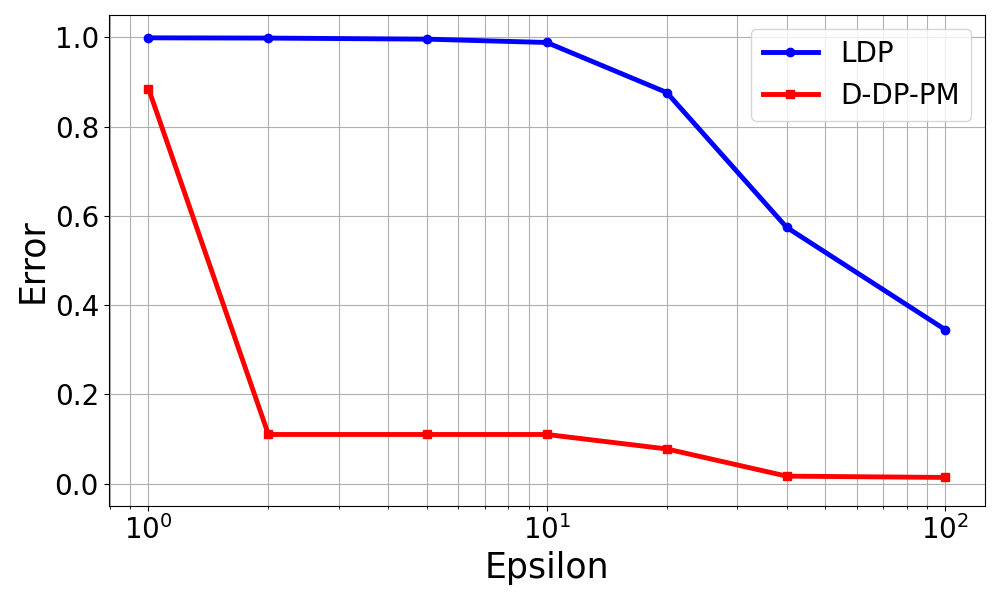}\label{fig:topright}}

\vspace{0.5em}

\subfloat[]{\includegraphics[width=0.42\textwidth]{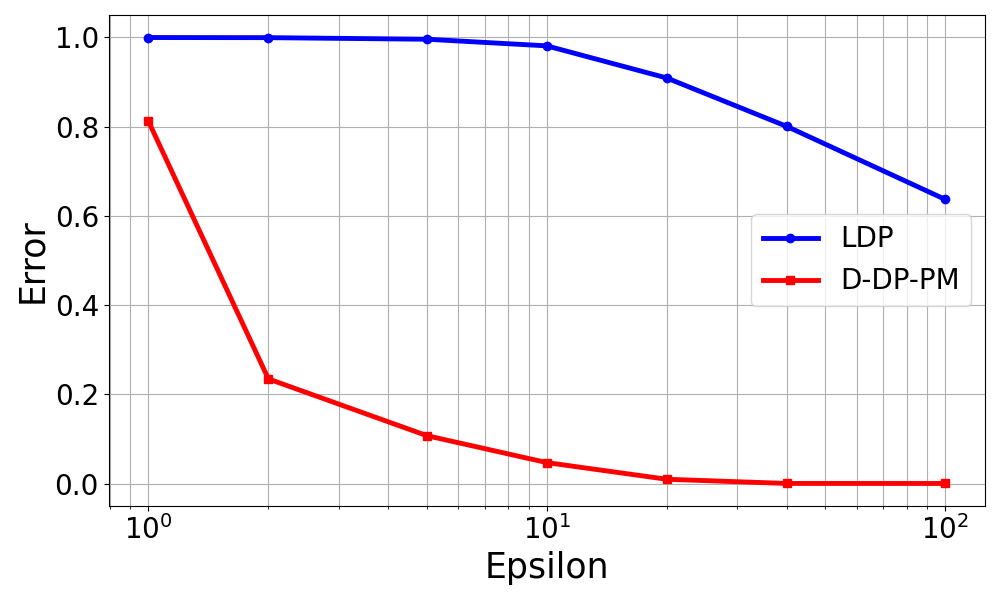}\label{fig:bottomleft}}
\hfill
\subfloat[]{\includegraphics[width=0.42\textwidth]{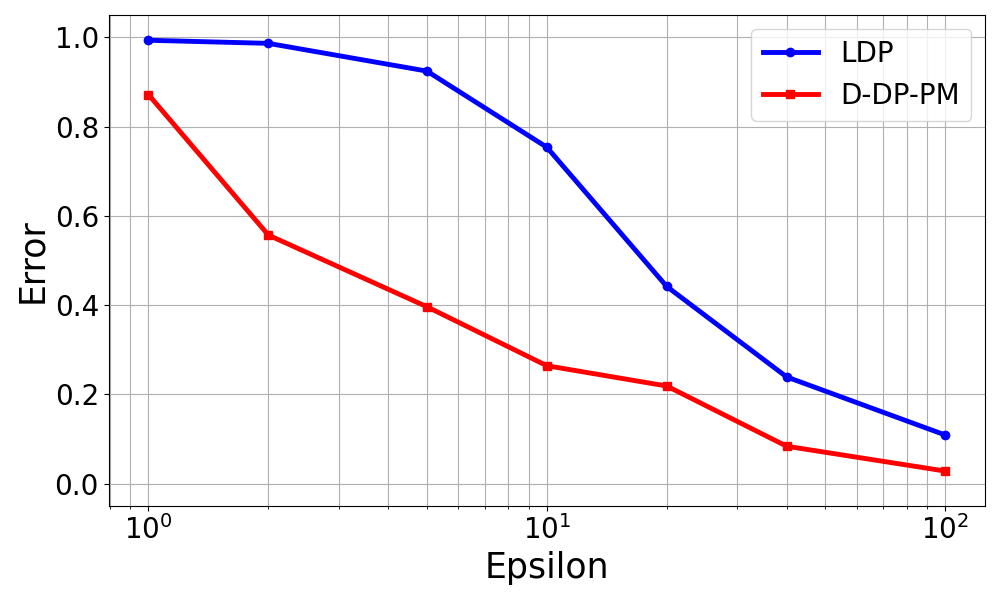}\label{fig:bottomright}}

\caption{All plots compare the performance of the D-DP-PM algorithm with that of adding noise directly to the distributed data. Error is measured using $\norm{\left(\mI-\vv\vv^\top\right)\vq}$. Top-left: Diabetes dataset \cite{efron2004least}. Top-right: Breast Cancer dataset \cite{breast_cancer_wisconsin_(diagnostic)_17}. Bottom-left: AMI dataset. Bottom-right: Wine dataset \cite{wine_109}.}
\label{fig:comparison}
\end{figure*}
\subsection{LDP-Method}
Formally, the LDP method adds Gaussian noise to each dataset $\mX_i$ such that the noise is distributed by $$\mG_i\sim\left(\vzero, \dfrac{2\ln(1.25/\delta)}{\epsilon^2}\mI\right).$$ That is, each agent constructs $\tilde{\mX}_i=\mX_i+\mG_i$ and shares $\tilde{\mX}_i$ with the entire network. Now each agent has a noisy version of the entire dataset $\mX$ and can run SVD to recover the eigenvectors. In implementation this was accomplished by adding noise directly to $\mX$ and doing an SVD on $\tilde{\mX}$.
\subsection{Privacy vs. Accuracy}
In order to evaluate the efficacy of the proposed method we implemented Algorithm \ref{alg:cov_approx_power} and tested it on numerous real datasets. The datasets of interest are the Diabetes dataset \cite{efron2004least}, the Wine dataset \cite{wine_109}, the Breast Cancer dataset \cite{breast_cancer_wisconsin_(diagnostic)_17}, and an Advanced Metering Infrastructure (AMI) dataset. Each dataset was preprocessed so that every point sat within, or on the surface of, the unit ball. This normalization was done to provide ``unit" sensitivity when constructing adjacent datasets $\mX'$. 

For each dataset we assumed a total of $4$ agents using the following mixing matrix representing a ring communication topology.
\begin{align*}
    \mW=\begin{bmatrix}
        .5 & .25 & 0 & .25\\
        .25 & .5 & .25 & 0\\
        0 & .25 &.5 &.25\\
        .25 & 0 & .25 & .5
    \end{bmatrix}
\end{align*}
To generate the distributed data, the raw global dataset $\mX$ was split row wise, according to \eqref{eq:our_data_dist}, into 4 equal parts with each agent receiving one of the parts. It is not an analytical requirement that each agent have the same number of samples, but it simplifies the implementation and was thus performed for practical reasons. 

To determine the appropriate choice of parameters, i.e. $\sigma^2_{\vp}(t)$, $T$, and $\alpha$ where chosen by initializing $\sigma^2_{\vp}(t)$ and $\alpha$ according to Corollary \ref{cor:params} followed by a grid search of near by values. That is, Monte-Carlo simulations were run over all the near by parameters for a fixed $\epsilon$ and $T$. The parameters that minimized error and privacy leakage were then chosen. In order to find the best choice of $T$ given the specified $\epsilon$, $T^*$ was computed in an offline fashion where $T^*$ was the number of power iterations required to achieve a convergence error of $10^{-3}$  without the presence of DP noise. Then, all other parameters were fixed and $T$ was varied around $T^*$. By performing this search over near by parameters a good choice of $\sigma^2_{\vp}(t), \alpha,$ and $T$ were found to minimize privacy leakage and error for a particular $\epsilon$. 

Given that the R\'enyi Divergence of Gaussians with different distributions is sensitive to rank deficient matrices, we apply a dimensionality reduction technique to the distributions of the releases $\vy^{(T)}_i$ to ensure the divergence can be computed. Formally, if we let 
\begin{align}
    \vy_i^{(T)}\sim \gN(\bm{\mu},\Sigma_{\vy}),
\end{align} then let $\mU^{(r)}$ denote the top $r$ eigenvectors. Then we can denote the full rank version of $\vy_i^{(T)}$ by $\vy_i^{(T)}(r)$ where 
\begin{align}
    \vy_i^{(T)}(r)\sim\gN\left((\mU^{(r)})^\top\bm{\mu},(\mU^{(r)})^\top\Sigma_{\vy}(\mU^{(r)})\right).
\end{align}
We apply this procedure prior to computing the R\'enyi divergence to ensure numerical stability.

Once the optimal parameters were found, an $\epsilon$ schedule and $\delta$ schedule were determined. Then each algorithm, the D-DP-PM algorithm and the LDP ran their perspective methods to approximate the eigenvector of $\mX\mX^\top$. For each $(\epsilon,\delta)$ pair, Monte-Carlo simulations were performed to more accurately approximate the error. All datasets used an $\epsilon$ schedule of $\begin{bmatrix}
    1 & 2 & 5 & 10 & 20 & 40 & 100 
\end{bmatrix}$. For the Diabetes data set, the upper limits on $\delta $ are given by $\begin{bmatrix}
    .38&.143&.008&2\cdot10^{-4}&2\cdot10^{-4}&10^{-4}&10^{-7}
\end{bmatrix}$. For the AMI dataset the $\delta$ limits are 
$\begin{bmatrix}
    .37 & .14 &.007 &5\cdot 10^{-5} &2\cdot 10^{-8} &4\cdot 10^{-17} & 4\cdot 10^{-43}
\end{bmatrix}$, while for Breast Cancer and Wine, respectively, we have
$\begin{bmatrix}
    .38 & .15 & .008 & .0002 & .0002 & 10^{-4} & 10^{-7}
\end{bmatrix}$, and 
$\begin{bmatrix}
    .44& .20 & .04 & .007 & .007 &.006 &.002
\end{bmatrix}$.
The results of the Monte-Carlo simulations given the $(\epsilon,\delta)$ constraints as above are presented in Figure \ref{fig:comparison}.

From Figure \ref{fig:comparison} we can see that the D-DP-PM algorithm has a substantial improvement in the privacy utility trade off when compared to the naive LDP method. This is especially clear in the moderate privacy setting where $\epsilon \in [2,5]$. One of the major advantages of the D-DP-PM algorithm, and the power method in general, is that it converges quickly. This means, for a small loss in accuracy, the algorithm can save on privacy by running fewer iterations.  Additionally, an interesting observation is that the D-DP-PM method performs better under higher dimensional data than lower dimensional. The AMI dataset contains 1352 samples with 25 dimensions while the Breast Cancer dataset contains 569 samples with 30 dimensions. 
\begin{figure*}
    \centering
    \includegraphics[width=0.42\linewidth]{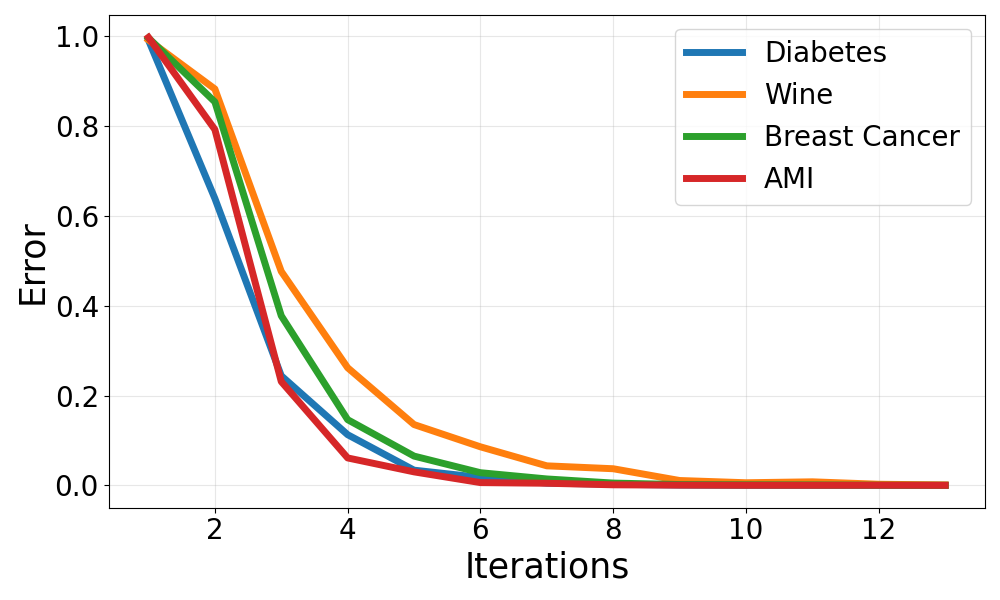}
    \includegraphics[width=.42\linewidth]{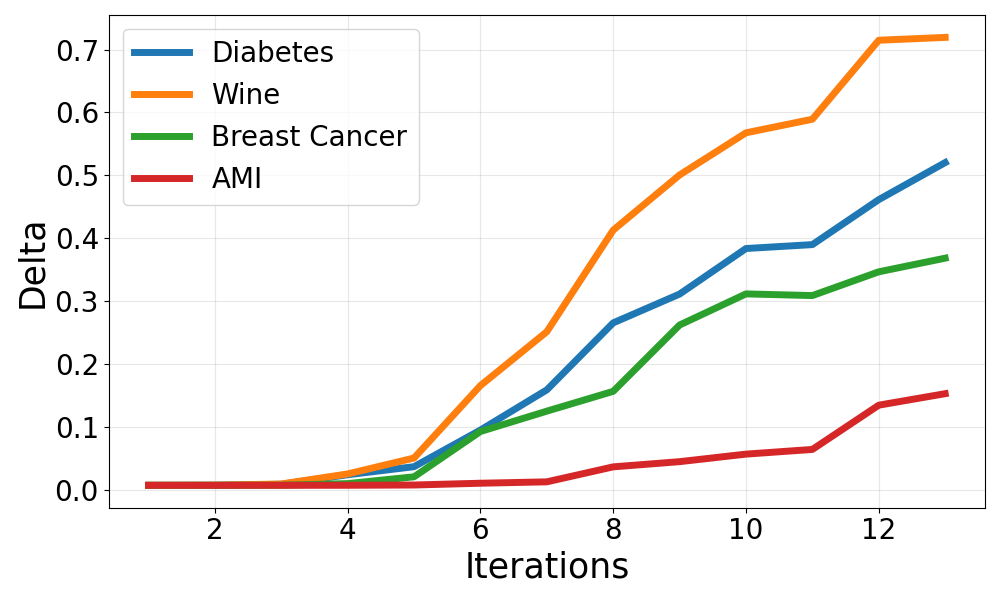}
    \caption{Left: For a fixed noise schedule, parameters and with $\epsilon=5$, the error as a function of the number of total iterations. Right: The corresponding $\delta$.}
    \label{fig:conv_delta}
\end{figure*}
\subsection{Convergence and Privacy bound}
We include an additional experiment where we fix $\epsilon=5$ for all data sets and fix the found parameters from the previous experiments for each dataset and plot the error and $\delta$ as a function of the number of iterations. See Figure \ref{fig:conv_delta}. The Figure provides further justification for the observation about how the D-DP-PM is advantageous since it converges quickly and thus can exchange iterations for higher privacy. This is clear in Figure \ref{fig:conv_delta} since by iteration 5 most datasets have converged and simultaneously their $\delta$ has only marginally increased.    
\section{Conclusion}
\label{sec:conclusion}
In this paper, we presented D-DP-PM, the first decentralized differentially private power method for eigenspace estimation when data is distributed row-wise across agents. Our approach addresses a fundamental gap in the literature by enabling privacy-preserving PCA in fully decentralized settings without central aggregation, while handling the challenging case where no single agent has access to the full dimensional space.
One of our key contributions is a novel algorithm design that preserves Gaussianity throughout iterations, enabling tractable privacy analysis through closed-form R\'enyi divergence bounds. Furthermore, we developed convergence analysis from first principles that explicitly captures the interplay between consensus errors and differential privacy noise. Furthermore, our empirical evaluation demonstrates that D-DP-PM substantially outperforms naive approaches across multiple datasets, with the advantage being most pronounced in moderate privacy settings. Additionally we show that privacy practitioners can exchange number of iterations for improved privacy.  

\appendix
\section{Appendix}
\subsection{Consensus Proofs}
\subsubsection{Consensus Error}
\label{proof:cons}
We claim that:
\begin{align}
\mX^\top\vq+\ve&=m\sum_{j=1}^m\left(\mW^c\right)_{ij}\mX_j^\top\vq_j \\
\norm{\ve}&=\gO\left(\lambda^c_2\left(\mW\right)\right)\nonumber.
\end{align}
\begin{proof}
    We drop the time indexing for clarity. We have that where $\vz=\begin{bmatrix}
        \vz_1 \\ \vdots \\ \vz_m
    \end{bmatrix}$, where $\vz_i=\mX_i^\top\vq_i$. Then we observe that \begin{align}
        \left(\left(\mW^c\otimes \mI_d\right)\vz)\right)_i=\sum_{j=1}^m\left(\mW^c\right)_{ij}\mX_j^\top\vq_j. 
    \end{align} Therefore if we let $\bar{\vz}:=\dfrac{\vone\vone^\top}{m}\vz$, then, because $\left(\mW^c\otimes \mI_d\right)\bar{\vz}=\bar{\vz}$   
    \begin{align}
        r(\vz):=\norm{\left(\left(\mW^c\otimes \mI_d\right)\vz)\right)-\bar{\vz}}&\leq \norm{\left(\mW^c\otimes \mI_d\right)(\vz-\bar{\vz})}\nonumber\\
        &\leq\lambda_2^{c}(\mW)\norm{\vz-\bar{\vz}}.\nonumber
    \end{align} 
    But since we are interested in $\mX^\top\vq$, and thus the sum of each $\vz_i$, we have an additional factor of $m$ meaning that, where $\ve_i$ denotes the $i$-th basis vector, 
    \begin{align}
        &\norm{\mX^\top\vq-m\sum_{j=1}^m\left(\mW^c\right)_{ij}\mX_j^\top\vq_j}\\
        &~~~~=\norm{m\left(\ve_i^\top\otimes \mI_d \right)r(\vz)}\leq m \lambda_2^{c}(\mW)\norm{\vz-\bar{\vz}}
    \end{align}
\end{proof}
\subsection{Privacy Proofs}
\subsubsection{$\delta$ bound}
\label{proof:dp}
\begin{proof}
    For shorthand let 
    \begin{align}
        L_{\mX\mX'}(\vy_i^{(T)}):=\ln\frac{f\left(\vy^{(T)}_i|\mX,\vq_i,\vp_i^{(1)},\hdots,\vp_i^{(T)}\right)}{f\left(\vy_i^{(T)}|\mX',\vq_i,\vp_i^{(1)},\hdots,\vp_i^{(T)}\right)}.
    \end{align} Then, the goal is to bound 
    \begin{align}
    \Pr\left(L_{\mX\mX'}>\epsilon\right).    
    \end{align}
    By direct application of the Chernoff bound we have
    \begin{align}
        &\Pr\left(L_{\mX\mX'}(\vy_i^{(T)})>\epsilon\right)\nonumber \\&~~~~\leq\inf_{\lambda}e^{-\lambda\epsilon}\E\left[e^{\lambda L_{\mX\mX'}(\vy_i^{(T)})}\right]\nonumber \\
        &~~~~=\inf_{\lambda} e^{-\lambda\epsilon}\int e^{\lambda L_{\mX\mX'}(\vy_i^{(T)})} f_{\vy_i^{(T)}|\mX,\vq_i,\vp_i^{(1)},\hdots,\vp_i^{(T)}}(\vy_i^{(T)}) d\vy_i^{(T)}.\label{eq:loglik_int}
    \end{align}
    By letting 
    \begin{align}
        P\sim p(\vy_i^{(T)})&:=f_{\vy_i^{(T)}|\mX,\vq_i,\vp_i^{(1)},\hdots,\vp_i^{(T)}}(\vy_i^{(T)}) d\vy_i^{(T)}\nonumber\\ 
        Q\sim q(\vy_i^{(T)})&:=f_{\vy_i^{(T)}|\mX',\vq_i,\vp_i^{(1)},\hdots,\vp_i^{(T)}}(\vy_i^{(T)}) d\vy_i^{(T)},
    \end{align}
    we can simplify \eqref{eq:loglik_int} to 
    \begin{align}
        \eqref{eq:loglik_int}&= \inf_{\lambda}e^{-\lambda\epsilon}\int p^{\lambda+1}(\vy_i^{(T)})q^{-\lambda}(\vy_i^{(T)}) d\vy_i^{(T)}.\label{eq:simp}
    \end{align}
    Finally we note that 
    \begin{align}
        \operatorname{D}_{\lambda+1}(P||Q)=\frac{1}{\lambda}\log\left(\int p^{\lambda+1}(\vy_i^{(T)})q^{-\lambda}(\vy_i^{(T)})\right). \label{eq:RDP}
    \end{align}
    By combining \eqref{eq:RDP} with \eqref{eq:simp} we have that 
    \begin{align}
        \eqref{eq:loglik_int}&=\inf_{\lambda} \exp\left\{-\lambda\epsilon +\lambda \operatorname{D}_{\lambda+1}(P||Q)\right\},
    \end{align}
    which completes the proof.
\end{proof}
\subsection{Convergence}
\subsubsection{Theorem \ref{th:conv}}
\begin{proof}
    \label{proof:conv}
    Let $\vv$ represent the principal eigenvector of $\mX\mX^\top$ and let $\tilde{\vq}:=\vq^{(T)}$ prior to normalization. Our goal is to bound 
\begin{align}
    R:=\dfrac{\norm{(\mI-\vv\vv^\top)\tilde{\vq}}}{\norm{\tilde{\vq}}}.
\end{align}
We know that 
\begin{align}
    \vq^{(t+1)}&=\alpha \operatorname{D}(\mX)\left(m(\mW^c\otimes \mI_d)\right)\operatorname{D}^\top(\mX)\vq^{(t)}+\vp^{(t)}\\
    \bm{\Xi}&:=\operatorname{D}(\mX)\left(m(\mW^c\otimes \mI_d)\right)\operatorname{D}^\top(\mX),
\end{align} 
which implies that 
\begin{align}
\tilde{\vq}&=\sum_{i=1}^n\left(\underbrace{ (\alpha\mu_i)^{T}\beta_i+\sum_{k=1}^T(\alpha\mu_i)^{T-k}\omega_i(k)}_{\bm{\Gamma}_i}\right)\vu_i \\ &\beta_i\sim \gN\left(0,\sigma_{\vq}^2\right) ~\text{ and }~\omega_i(k)\sim\gN\left(0,\sigma_p(k)^2\right)\nonumber \\
\implies \nonumber \\
\tilde{\vq}&=\sum_{i=1}^n\Gamma_i\vu_i.\nonumber
\end{align}
Similarly, we can fully specify the distribution of $\tilde{\vq}$ by
\begin{align}
    \tilde{\vq}\sim \gN\left( \vzero, \underbrace{\alpha^{2T}\sigma^2_{\vq}\bm{\Xi\Xi}^\top+\sum_{k=1}^{T}\alpha^{2T-k}\sigma^2_{\vp}(k)\left(\bm{\Xi\Xi}^\top\right)^{T-k}}_{\bm{\Omega}}\right).
\end{align}
Furthermore, we denote $\vu$ as the principal eigenvector of $\bm{\Xi}$.
We define $\mP:=(\mI-\vv\vv^\top)$ and $\mQ:=\bm{\Omega}^{1/2}\mP\bm{\Omega}^{1/2}$. By doing so we can then write
\begin{align}
    \tilde{q}\sim \bm{\Omega}^{1/2}\vs \quad \vs\sim\gN(\vzero,\mI_n).
\end{align}
This conveniently yields 
\begin{align}
    \left(\dfrac{\norm{(\mI-\vv\vv^\top)\tilde{\vq}}}{\norm{\tilde{\vq}}}\right)^2=\dfrac{\vs^\top\mQ\vs}{\vs^\top\bm{\Omega}\vs}.
\end{align}
Therefore, we choose a constant $\Theta$ such that  
\begin{align}
    \left(\dfrac{\norm{(\mI-\vv\vv^\top)\tilde{\vq}}}{\norm{\tilde{\vq}}}\right)^2\leq \Theta,
\end{align}
yielding the classic quadratic equation 
\begin{align}
    \vs^\top\left(\mQ-\Theta\bm{\bm{\Omega}}\right)\vs\leq 0.
\end{align}
We then introduce another constant $\Delta > 0$ so that 
\begin{align}
    \Theta=1-\dfrac{\vv^\top\vv}{\Tr(\Omega)}+\Delta.
\end{align}
This further implies that 
\begin{align}
    \Pr\left(R^2\geq \Theta)=\Pr(\vs^\top\left(\mQ-\Theta\bm{\bm{\Omega}}\right)\vs>c\right) \label{eq:R_eq}
\end{align}
Therefore, we can apply the Hanson-Wright \cite{Vershynin_2018} tail bound to
\begin{align}
    \Pr\left(\vs^\top\left(\mQ-\Theta\bm{\bm{\Omega}}\right)\vs -\Tr( \vs^\top\left(\mQ-\Theta\bm{\bm{\Omega}}\right)\vs)\geq c \right)\leq \gamma.
\end{align}
We notice that 
\begin{align}
    \Tr( \vs^\top\left(\mQ-\Theta\bm{\bm{\Omega}}\right)\vs)&=\Tr(\mQ)-\Theta\Tr(\bm{\Omega})\nonumber\\
    &=\Tr(\bm{\Omega})-\vv^\top\bm{\Omega}\vv-\Theta\Tr(\bm{\Omega})\nonumber \\
    &=-\Delta\Tr(\bm{\Omega}).
\end{align}
There if we choose 
\begin{align}
    &\Delta \geq \nonumber \\ &~~\dfrac{2\norm{ \left(\mQ-\Theta\bm{\bm{\Omega}}\right)}_F\sqrt{\log(\frac{1}{\gamma})}+2\norm{ \left(\mQ-\Theta\bm{\bm{\Omega}}\right)}_2\log(\frac{1}{\gamma})}{\Tr(\bm{\Omega})},
\end{align}
we can apply Hanson-Wright \cite{Vershynin_2018} and we get 
\begin{align}
    &\Pr\left( \vs^\top\left(\mQ-\Theta\bm{\bm{\Omega}}\right)\vs-\Tr( \vs^\top\left(\mQ-\Theta\bm{\bm{\Omega}}\right)\vs)\right.\nonumber\\&~~~~~\left. \geq \norm{ \left(\mQ-\Theta\bm{\bm{\Omega}}\right)}_F\sqrt{\log(1/\gamma)})+\norm{ \left(\mQ-\Theta\bm{\bm{\Omega}}\right)}_{2}\log(1/\gamma)\right)\nonumber\\
    &~~~~~~~~~~\leq \gamma.
\end{align}
Finally, plugging this back into \ref{eq:R_eq} we have that 
\begin{align}
    &\dfrac{\norm{(\mI-\vv\vv^\top)\tilde{\vq}}^2}{\norm{\tilde{\vq}}^2}\leq\nonumber\\&~~~\dfrac{2\norm{ \left(\mQ-\Theta\bm{\bm{\Omega}}\right)}_F\sqrt{\log(1/\gamma)}+2\norm{ \left(\mQ-\Theta\bm{\bm{\Omega}}\right)}_2\log(1/\gamma)}{\Tr(\bm{\Omega})}\nonumber \\
    &~~~~~~~+1-\dfrac{\vv^\top\bm{\Omega}\vv}{\Tr(\bm{\Omega})}\nonumber\\
    &\qquad\text{w.p.  } 1-\gamma
\end{align}
In order to see the convergence in terms of spectrum of $\mX\mX^\top$ and $\bm{\Xi}$ we need to bound 
\begin{align}
    1-\dfrac{\vv^\top\bm{\Omega}\vv}{\Tr(\bm{\Omega})}. \label{eq_tr_bound}
\end{align}
we let $\vu$ denote the principal eigenvector of $\bm{\Xi}$ and we assume that 
\begin{align}
    \norm{\bm{\Xi}-\mX\mX^\top}_2\leq E \qquad E\leq \lambda_1(\mX\mX^\top)-\lambda_2(\mX\mX^\top).
\end{align}
The fundamental idea is that since, $ \E\left[\sin^2(\vv,\tilde{\vq})\right]=1-\dfrac{\vv^\top\bm{\Omega}\vv}{\Tr\left(\bm{\Omega}\right)} $, 
we can bound \eqref{eq_tr_bound} by 
\begin{align}
    \sin^2(\vv,\tilde{\vq})\leq 2\sin^2(\vv,\vu)+2\sin^2(\vu,\tilde{\vq}).
\end{align} Using this structure we first start with $\sin^2(\vu,\tilde{\vq})$.
By Davis-Kahan\cite{davis1970rotation}, we have that 
\begin{align}
    \sin\left(\vu,\vv\right)&\leq \dfrac{\norm{\bm{\Xi}-\mX\mX^{\top}}_2}{\lambda_1(\mX\mX^\top)-\lambda_2(\mX\mX^\top)}\nonumber\\&\leq \dfrac{E}{\lambda_1(\mX\mX^\top)-\lambda_2(\mX\mX^\top)}
\end{align}
Now, if we can bound the ratio 
\begin{align}
    \dfrac{\Var\left[\bm{\Gamma}_i\right]}{\Var\left[\bm{\Gamma}_1\right]},
\end{align}
we can bound 
\begin{align}
    1-\dfrac{\vu_1^\top\bm{\Omega}\vu_1}{\Tr\left(\bm{\Omega}\right)}
\end{align} 
since, by the fact that $\Tr(\bm{\Omega})=\sum_{i=1}^n\Var\left[\bm{\Gamma}_i\right]$ and $\vu_1^\top\bm{\Omega}\vu_1=\Var\left[\bm{\Gamma}_i\right]$ (by orthogonality of $\vu$),
\begin{align}
    1-\dfrac{\vu_1^\top\bm{\Omega}\vu_1}{\Tr\left(\bm{\Omega}\right)}=\dfrac{\sum_{i=2}^n\Var[\bm{\Gamma}_i]}{\sum_{i=1}^n\Var\left[\bm{\Gamma}_i\right]}\leq (n-1)\dfrac{\Var\left[\bm{\Gamma}_i\right]}{\Var\left[\bm{\Gamma}_1\right]}.
\end{align}
First notice that $\Var\left[\bm{\Gamma}_i\right]=(\alpha\mu_i)^{2T}\sigma_\vq^2+\sum_{k=1}^{T}(\alpha\mu_i)^{2(T-k)}\sigma_\vp^2(k)$, then we have 
\begin{align}
    \dfrac{\Var\left[\bm{\Gamma}_i\right]}{\Var\left[\bm{\Gamma}_1\right]}&=\dfrac{(\alpha\mu_i)^{2T}\sigma_\vq^2+\sum_{k=1}^{T}(\alpha\mu_i)^{2(T-k)}\sigma_\vp^2(k)}{(\alpha\mu_1)^{2T}\sigma_\vq^2+\sum_{k=1}^{T}(\alpha\mu_1)^{2(T-k)}\sigma_\vp^2(k)}\nonumber \\
    &\leq \left(\dfrac{\mu_2}{\mu_1}\right)^{2T}\dfrac{\sigma_\vq^2+\sum_{k=1}^{T}(\alpha\mu_i)^{2(T-k)}(\alpha\mu_2)^{-2T}\sigma_\vp^2(k)}{\sigma_\vq^2+\sum_{k=1}^{T}(\alpha\mu_1)^{2(T-k)}(\alpha\mu_1)^{-2T}\sigma_\vp^2(k)}\nonumber\\
    &~~~~~~\text{ letting }\rho:=\left(\dfrac{\sigma_\vq^2+\sum_{k=1}^{T}(\alpha\mu_2)^{-2k}\sigma_\vp^2(k)}{\sigma_\vq^2+\sum_{k=1}^{T}(\alpha\mu_1)^{-2k}\sigma_\vp^2(k)}\right)\nonumber \\
    &=\left(\dfrac{\mu_2}{\mu_1}\right)^{2T}\rho.
\end{align}
Therefore, we have that 
\begin{align}
    1-\dfrac{\vv^\top\bm{\Omega}\vv}{\Tr(\bm{\Omega})}\leq 2\left(\dfrac{E}{\lambda_1-\lambda_2}+(n-1)\rho\left(\dfrac{\mu_2}{\mu_1}\right)^{2T}\right).
\end{align}
Furthermore, we have that $E\leq n_im\lambda_2^c(\mW)$. Therefore our final bound becomes  \begin{align}
    &\dfrac{\norm{(\mI-\vv\vv^\top)\tilde{\vq}}^2}{\norm{\tilde{\vq}}^2}\leq\nonumber \\
      &~~~~~\dfrac{2\norm{ \left(\mQ-\Theta\bm{\bm{\Omega}}\right)}_F\sqrt{\log(1/\gamma)}+2\norm{ \left(\mQ-\Theta\bm{\bm{\Omega}}\right)}_2\log(1/\gamma)}{\Tr(\bm{\Omega})} \nonumber\\
      &~~~~~+2\left(\dfrac{n_im\lambda_2^c(\mW)}{\lambda_1-\lambda_2}+(n-1)\rho\left(\dfrac{\mu_2}{\mu_1}\right)^{2T}\right)\nonumber \\
    &\qquad\text{w.p  } 1-\gamma,
\end{align}
thus completing the proof.
\end{proof}
\subsubsection{Corollary \ref{cor:params}}
\begin{proof}
\label{proof:params}
    \begin{align}
    &\left(\dfrac{\sigma_\vq^2+\sum_{k=1}^{T}(\alpha\mu_2)^{-2k}\sigma_\vp^2(k)}{\sigma_\vq^2+\sum_{k=1}^{T}(\alpha\mu_1)^{-2k}\sigma_\vp^2(k)}\right)\nonumber \\ 
    &~~~\leq \left(\dfrac{\sigma_\vq^2+\sum_{k=1}^{T}(\alpha\mu_2)^{-2k}\sigma_\vp^2(k)}{\sigma^2_\vq}\right)\nonumber \\
    &~~~\leq \left(\dfrac{\sigma_\vq^2+\sum_{k=1}^{T}(\alpha\mu_2)^{-2k}\left(\dfrac{\mu_2}{\mu_1}\right)^{2k}}{\sigma^2_\vq}\right)\text{ by } \sigma_{\vp}(t)\leq \left(\dfrac{\mu_2}{\mu_1}\right)^{t}\nonumber\\
    &~~~=\left(\dfrac{\sigma_\vq^2+\sum_{k=1}^{T}(\alpha\mu_1)^{-2k}}{\sigma^2_\vq}\right).\label{eq_param_final}
    \end{align}
    Since $\alpha\mu_1>1$ we can then upper bound \eqref{eq_param_final} by its geometric series yielding 
    \begin{align}
        \eqref{eq_param_final}\leq 1+ \dfrac{1}{\sigma^2_{\vq}(1-(\alpha\mu_1)^{-2})},
    \end{align}
    thus completing the proof.
\end{proof}
\bibliographystyle{IEEEtran}
\bibliography{ref}

\end{document}